%% file: acl_latex.tex
\pdfoutput=1

\documentclass[11pt]{article}

\usepackage[]{acl}

\usepackage{times}
\usepackage{latexsym}
\usepackage{subcaption}
\usepackage{multirow}
\usepackage{float}
\usepackage{hyperref}
\usepackage{amsmath,amssymb}

\usepackage[T1]{fontenc}

\usepackage[utf8]{inputenc}

\usepackage{microtype}
\usepackage{todonotes}

\definecolor{green}{rgb}{0,0.6,0}

\title{Faithful or Extractive? On Mitigating the Faithfulness-Abstractiveness Trade-off in Abstractive Summarization}

\author{
Faisal Ladhak$^1$\Thanks{ Equal contribution. Corresponding author for queries: \href{mailto:faisal@cs.columbia.edu}{faisal@cs.columbia.edu}.}~\;~ Esin Durmus$^2$\footnotemark[1]~\;~ He He$^3$~\;~ Claire Cardie$^4$~\;~Kathleen McKeown$^1$\\
$^1$Columbia University~\;~ $^2$Stanford University~\;~\\$^3$New York University~\;~$^4$Cornell University
}
\begin{document}
\maketitle

\input{abstract}

\begin{figure*}[t]
\centering
\includegraphics[width=0.86\linewidth]{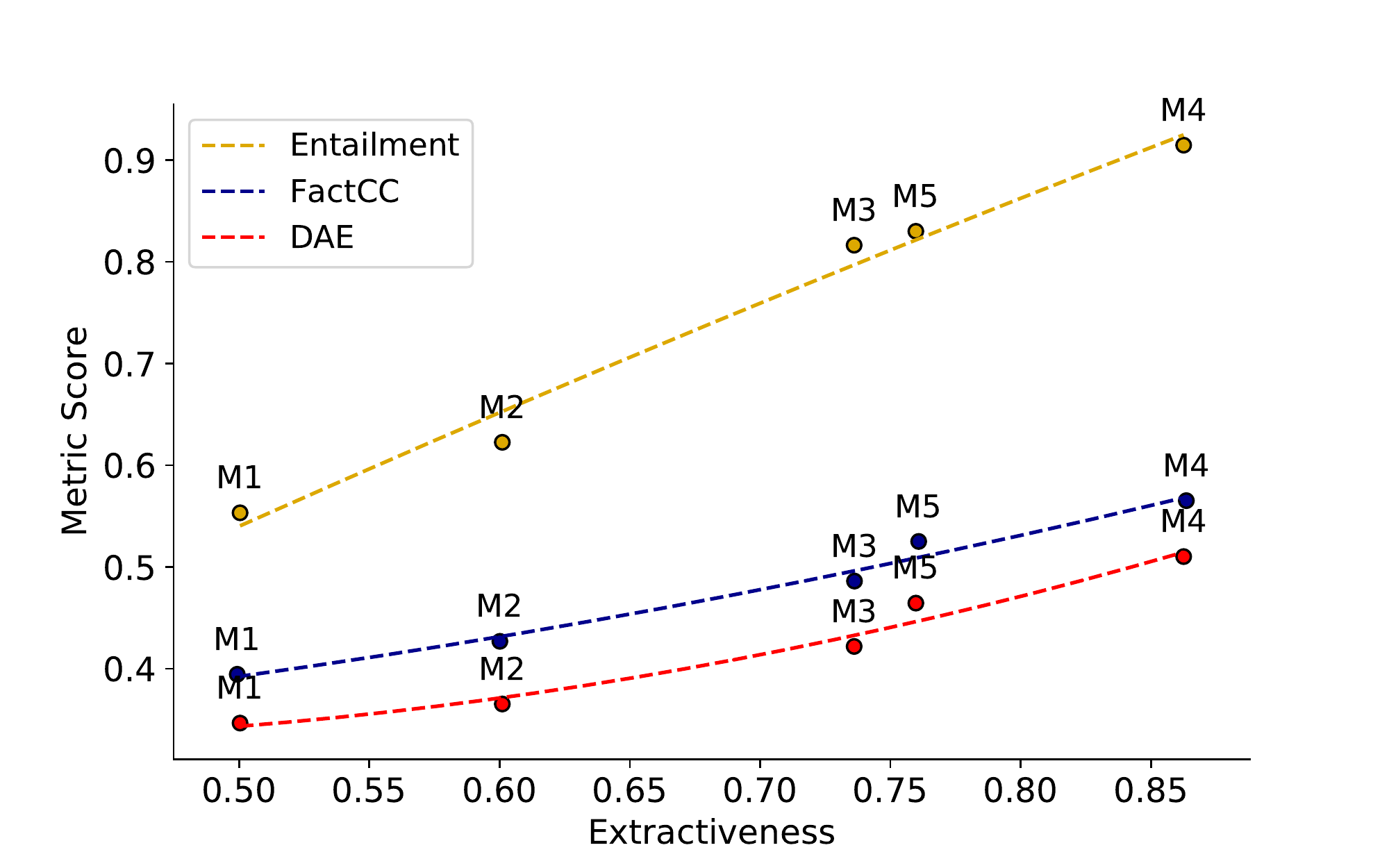}
\caption{Extractiveness of generated outputs versus automated metric scores for Entailment, FactCC and DAE on the Gigaword dataset. We use \textit{coverage} defined in \newcite{grusky-etal-2018-newsroom} to measure extractiveness, where summaries with higher coverage are more extractive. We observe that automated metrics of faithfulness are positively correlated with extractiveness.}
\label{fig:density_metrics}

\end{figure*}

\input{intro}

\input{dataset}

\input{analysis}

\input{effective_faithfulness}

\input{selector}

\input{related_work}

\input{implications}

\bibliography{anthology,custom}
\bibliographystyle{acl_natbib}

\begin{table*}[h]
\begin{center}
\begin{tabular}{|c|lrrr|}
\hline
 Dataset & Quartile & \# Examples  &  Article Length & Summary Length  \\ 
    \hline
    \multirow{4}{*}{Gigaword} 
    & Q1  & 985,931 & 30.58 & 8.03 \\
    & Q2  & 961,970 & 32.02 & 8.32 \\ 
    & Q3  & 952,833 & 31.77 & 8.41 \\ 
    & Q4  & 903,223 & 31.05 & 8.17 \\ 
    \hline
    
    \multirow{4}{*}{Wikihow} 
    & Q1  & 328,470 & 50.73 & 7.63 \\
    & Q2  & 221,452 & 75.69 & 7.40 \\ 
    & Q3  & 206,558 & 85.44 & 5.96 \\ 
    & Q4  & 243,837 & 92.09 & 5.49 \\ 
    \hline
    
\end{tabular}
\end{center}
\caption{Data statistics for each quartile. \textit{Length} corresponds to average \# of words.}
\label{tab:data_stats}
\end{table*}

\begin{figure*}[]
\centering
\includegraphics[width=\linewidth]{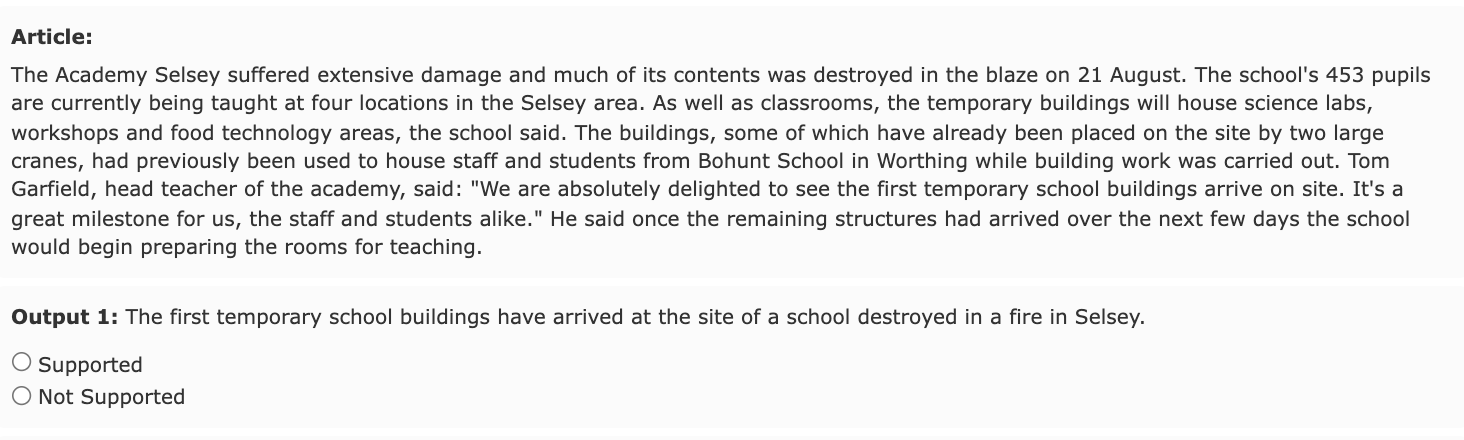}
\caption{An example from our human evaluation.}
\label{fig:example_human_eval}

\end{figure*}

\newpage
\appendix
\input{appendix}

\end{document}

%% file: abstract.tex
\begin{abstract}
Despite recent progress in abstractive summarization, systems still suffer from faithfulness errors. 
While prior work has proposed models that improve faithfulness, it is unclear whether the improvement comes from an increased level of extractiveness of the model outputs as one naive way to improve faithfulness is to make summarization models more extractive.
In this work, we present a framework for evaluating the \textit{effective faithfulness} of summarization systems, by generating a \textit{faithfulness-abstractiveness trade-off curve} that serves as a \textit{control} at different operating points on the abstractiveness spectrum. 
We then show that the baseline system as well as recently proposed methods for improving faithfulness, fail to consistently improve over the \textit{control} at the same level of abstractiveness.
Finally, we learn a selector to identify the most faithful and abstractive summary for a given document, and show that this system can attain higher faithfulness scores in human evaluations while being more abstractive than the baseline system on two datasets. Moreover, we show that our system is able to achieve a better faithfulness-abstractiveness trade-off than the \textit{control} at the same level of abstractiveness.

\end{abstract}

%% file: intro.tex
\section{Introduction}

Generating abstractive summaries of documents has been a long-standing goal of summarization. While there has been tremendous progress towards this goal \cite{kryscinski-etal-2018-improving, dong2019unified, zhang2019pegasus, lewis-etal-2020-bart}, abstractive summarization systems still suffer from faithfulness errors \cite{CaoWeiLiLi2018}, generating information that is not present in the original text. This has led to an increased research in faithfulness evaluation of summarization systems \cite{falke-etal-2019-ranking, kryscinski-etal-2020-evaluating, durmus-etal-2020-feqa} as well as methods to improve faithfulness of generated summaries \cite{kang-hashimoto-2020-improved, chen-etal-2021-improving}. 
Intuitively, one straightforward way of improving faithfulness of generated summaries is to copy a larger amount of content from the source article (i.e.\ more extraction). %
Thus, any methods that increase the level of extractiveness, whether intentionally or 
not, would improve faithfulness.
Without reported extractiveness, it is unclear whether prior improvements mainly arise from increased extractiveness.
We argue that in order to make progress in abstractive summarization, it is important to tease apart faithfulness improvements due to increased extractiveness 
versus improvements due to improved abstraction. 

In order to tease this apart, we develop a framework for evaluating progress in faithfulness, by considering the \textit{effective faithfulness}, i.e. the improvement in faithfulness over a baseline system (\textit{control}) operating at the same level of extractiveness. In particular, we split the training examples into different groups by the extractiveness of the summary, and train the \textit{control models} on each group.
Each of these models corresponds to a specific tradeoff between abstractiveness and faithfulness,
forming a \textit{trade-off curve}
indicating how much faithfulness can be improved solely by increasing extractiveness. Systems that improve \textit{effective faithfulness} should lie above this curve. 

Using this framework,
we show that the improved faithfulness of recently proposed methods comes mainly from an increased extractiveness. We then conduct further analysis to explore whether it is possible to have a system that can be both more abstractive and more faithful than the baseline system. We train a selector on a small set of human-annotated data that, given a set of output summaries with varying levels of extractiveness, picks the most abstractive output that is faithful to the source.
Our proposed system is both more abstractive and more faithful than the baseline.
Moreover, we show that our system is able to improve the \textit{effective faithfulness}, achieving a better trade-off than the \textit{control} at the same point on the abstractiveness spectrum.

To summarize, our contributions are as follows:

\begin{enumerate}
    \item We present a framework to evaluate the progress in improving 
    \textit{effective faithfulness} of models considering the \textit{control} at the same level of extractiveness. 
    \item We illustrate the importance of considering \textit{effective faithfulness} by showing that recently proposed methods for improving faithfulness are able to attain higher faithfulness scores than the baseline, but do not consistently improve over the \textit{control curve}, indicating that most of their improvements come from generating more extractive outputs, on average.
    \item We propose a selector that picks the most abstractive and faithful summary from a set of possible summaries, and show that this method gets higher effective faithfulness compared to the existing methods. 
\end{enumerate}

%% file: dataset.tex
\section{Dataset}
We conduct our study 
on
two English abstractive summarization datasets, one from the news domain, and one from a non-news domain. For the news domain dataset, we decided against using the popular CNN/Dailymail dataset since its reference summaries tend to be very extractive \cite{kedzie-etal-2018-content, bommasani-cardie-2020-intrinsic}, making it a poor choice for studying faithfulness in abstractive summarization. Similarly, we also decided against using XSum, another popular news summarization dataset, since almost $77\%$ of the gold reference summaries contain hallucinations \cite{maynez-etal-2020-faithfulness}. Instead, we opted for Gigaword and Wikihow, which are datasets with substantial abstraction without as much hallucination problems as XSum. Gigaword reference summaries have substantially less hallucinations than XSum \cite{kang-hashimoto-2020-improved}, and WikiHow summaries tend to be of a higher quality since they are written and curated by humans \cite{koupaee2018wikihow, ladhak-etal-2020-wikilingua}.

\noindent\textbf{Wikihow} \cite{koupaee2018wikihow} is a dataset of how-to articles covering a diverse set of topics, collected from the wikihow.com website. Each article contains several paragraphs detailing step by step instructions for a procedural task. There are about $12$M such paragraphs in the dataset, paired with a one sentence summary. 

\noindent\textbf{Gigaword} \cite{rush_2015} is a headline generation dataset that contains around $4$M examples, extracted from news articles that were collected as part of the Gigaword corpus \cite{graff2003english}. The model is tasked with generating the headline of the article given the first sentence.

\subsection{Dataset Extractiveness}
We follow the process detailed by \newcite{grusky-etal-2018-newsroom}, and use \textit{extractive fragment coverage} and \textit{extractive fragment density} as the measures of extractiveness of a given summary. Henceforth we will refer to these as coverage and density respectively. Coverage is the percentage of words in a summary that are from the source article. Density is the average length of the text spans copied from the document
that are contained in the summary. A summary that copies larger chunks of text from the source article will have a higher density.

%% file: analysis.tex
\section{Analysis on Metrics of Faithfulness}

Recent studies of faithfulness evaluation have proposed model-based automated metrics to detect whether a given summary is faithful to the source article. For example, \newcite{falke-etal-2019-ranking} \textbf{(Entailment)}
have
studied using pretrained entailment based methods to assess the probability of the generated output being entailed by the source article. \newcite{kryscinski-etal-2020-evaluating} \textbf{(FactCC)}
 augment hallucinated summaries 
by applying rule-based transformations to the document sentences and train a BERT-based model to classify whether the generated output is faithful. \newcite{goyal2020annotating} \textbf{(DAE)}
have collected fine-grained annotations to study word-, dependency- and sentence-level faithfulness and use these annotations to train a factuality detection model. 
 
Figure \ref{fig:density_metrics} 
shows the relationship between the 
average coverage of the generated outputs (extractiveness) vs. average metric scores (faithfulness) assigned to various abstractive summarization models trained on Gigaword.\footnote{These are the baseline and quartile models that are described in \S \ref{tradeoff}.}
We observe that there is a positive correlation 
between extractiveness and faithfulness scores, as models whose generated summaries have a higher average coverage 
tend to also get higher scores for each of the faithfulness
metrics. 
This correlation between exractiveness and faithfulness makes it unclear whether a model gets higher factuality scores simply because it is more extractive or it is capable of generating faithful summaries at the original level of extractiveness. This highlights the need for accounting for extractiveness in order to compare faithfulness across different abstractive summarization systems.

%% file: effective_faithfulness.tex
\section{Evaluating Effective Faithfulness}
Given that extractiveness is confounded with faithfulness, we propose a framework for evaluating \textit{effective faithfulness}, which takes into account the extractiveness of a system. In order to do this, we first need to determine the 
faithfulness of a system operating at a given level of extractiveness. We call this the \textit{Faithfulness-Abstractiveness Tradeoff} and we describe it further in \S\ref{tradeoff}. The \textit{effective faithfulness} of a system is then simply the relative difference between the faithfulness score assigned to the system, and the score of a system operating with the same average extractiveness according to the trade-off curve.

\subsection{Faithfulness-Abstractiveness Tradeoff}
\label{tradeoff}

In order to understand the effectiveness of a proposed system for improving faithfulness, we need to be able to account for its extractiveness. We finetune pre-trained BART models \cite{lewis-etal-2020-bart}
for different levels of extractiveness, without any explicit recourse for improving faithfulness. We then use these systems to create a \textit{faithfulness-abstractiveness trade-off curve} that can serve as a control to measure the \textit{effective faithfulness} of summarization systems. Models that improve \textit{effective faithfulness} should lie above the  \textit{faithfulness-abstractiveness trade-off curve}.\footnote{Human evaluation data and trade-off curves can be found at \href{https://github.com/fladhak/effective-faithfulness}{https://github.com/fladhak/effective-faithfulness}.}

\begin{table*}[t]
    \centering
    \begin{tabular}{|c|l|}
    \hline
   Article &  \begin{minipage}[t]{1.8\columnwidth}%
Once you decide what to outsource, look for the right contractors. Start by asking for referrals from your own professional network. Talk to other business owners and professionals about how and where they outsource. You can also check professional associations or trade groups field in which you are trying to outsource work. Use other social media platforms such as Facebook or Twitter to advertise what you are looking for. Alternately, you can connect with contractors and freelancers on sites such as eLance, Guru and oDesk. These websites allow business owners to place an ad that describes what kind of work they need to have done, and contractors respond with their qualifications and rates.  [TRUNCATED] ...
\end{minipage}\tabularnewline 

    \hline
    Baseline & {\color{green}Search} for contractors and freelancers to outsource the work. \\ 
    \hline
    Q1 & {\color{green}Conduct} an {\color{green}initial search} for {\color{green}qualified} contractors and freelancers.   \\ 
    \hline
    Q2 & {\color{green}Search} for {\color{green}qualified} contractors and freelancers to work on your {\color{green}project}.   \\ 
    \hline 
    Q3 &  {\color{green}Search} for contractors and freelancers to  {\color{green}do} the work.  \\ 
    \hline
    Q4 &  Look for contractors and freelancers to bid on the work.  \\
    \hline
    \end{tabular}
    \caption{Example summaries generated by the baseline and quartile models for the article ``\textit{How to Outsource Small Business Tasks}'' from Wikihow dataset. The tokens that do not appear in the source article are indicated by {\color{green}green}.}
    \label{tab:example_summaries}
\end{table*}

\begin{table}[h]
\begin{center}
\begin{tabular}{|c|lrr|}
\hline
 Dataset & Model  &  Coverage & Faithfulness  \\ 
    \hline
    \multirow{4}{*}{Gigaword} 
    &  Baseline  & 76.12  & 83.33 \\
    & Q1  & 50.25 & 71.83 \\
    & Q2  & 60.57 & 79.50 \\ 
    & Q3  & 73.64 & 86.67 \\ 
    & Q4  & 86.94 & 89.17 \\ 
    \hline
    
    \multirow{4}{*}{Wikihow} 
    &  Baseline &  88.28 & 82.52\\
    & Q1  & 81.34 & 67.82 \\
    & Q2  & 85.34 & 76.21 \\ 
    & Q3  & 87.59 & 80.35 \\ 
    & Q4  & 90.19 & 91.08\\ 
    \hline
    
\end{tabular}
\end{center}
\caption{Coverage and faithfulness values of the baseline and each quartile model for Gigaword and Wikihow. Quartile models with higher coverage have higher faithfulness scores.}
\label{tab:quartile_coverage_faithfulnesss}
\end{table}

In particular, we sub-sample the training data into extractiveness quartiles by computing the coverage of the references with respect to the source articles. We then fine-tune BART on each of these quartiles to obtain \textbf{quartile models} with varying level of extractiveness. In addition, we also finetune BART on all of the data, which we call the \textbf{baseline}.

We collect faithfulness annotations for summaries generated by each of these models for a random sample of $200$ articles. We collect three annotations per example on 
Amazon Mechanical Turk asking whether an output is faithful or unfaithful with respect to the corresponding source article. We then compute the percentage of annotators that selects "faithful", and use this as the faithfulness score for each example.\footnote{Details of the human evaluation are included in \autoref{human_eval_appendix}.}

Table \ref{tab:quartile_coverage_faithfulnesss} shows the coverage and faithfulness scores for the baseline and the quartile models, where Q1 is the \textit{most abstractive} and Q4 is the \textit{most extractive} quartile.\footnote{Additional dataset statistics are shown in \autoref{sec:appendix_data_stats}.} We observe that the models that are fine-tuned on more extractive quartiles produce outputs with significantly higher coverage and faithfulness scores. The baseline model generates relatively extractive outputs with coverage closest to Q3 on both Gigaword and Wikihow. Furthermore, we observe that the baseline model has a higher coverage than the model fine-tuned on Q3 but it has lower faithfulness score for Gigaword.

Table \ref{tab:example_summaries} shows an article from the Wikihow dataset and corresponding output summaries generated by the baseline and each of the quartile models. We observe that the generated summaries are very similar in meaning; however,  the output generated by the Q1 model includes a higher number of novel words (i.e. lower coverage) compared to the other models while staying faithful to the article. Conversely, Q4 model has a coverage of 1 in this example; all the words generated by this model are from the source article. On average, the Q1 model
generates outputs that are more abstractive and less faithful while Q4 generates outputs that are more extractive and more faithful.

%% file: selector.tex
\section{Mitigating the Trade-off}

\subsection{Oracle Experiments}
\label{oracle}
We first aim to understand whether it is possible to mitigate
the faithfulness-abstractiveness tradeoff by designing several oracle experiments where we have access to human judgments. 

\begin{table}[]
    \centering
    \begin{tabular}{|c|lrr|}
    \hline
    Dataset &   &  Cov. & Faithfulness  \\ 
    \hline
    \multirow{4}{*}{Gigaword} 
     &   Baseline   & 76.12 &  83.33  \\
     &   bf  & 77.74  &  89.57 \\
     &   bfe  & 61.87 &  90.67  \\ 
     &   qfe  & 63.55 &  98.00 \\ 
    \hline
    
    \multirow{4}{*}{Wikihow} 
     &   Baseline      & 82.52  & 88.28\\
     &   bf    & 83.95  & 92.20\\
     &   bfe  & 70.52  & 91.32 \\ 
     &   qfe & 72.58  &  98.61\\ 
    \hline
    
    \end{tabular}
    \caption{Oracle coverage and faithfulness values for Gigaword and Wikihow. The oracle analysis suggests that being able to control for extractiveness can allow us to build systems that mitigate the trade-off.}
    \label{tab:oracle_experiments}
\end{table}

\begin{table*}[t]
    \centering
    \begin{tabular}{|l|rr|rr|}
    \hline
     & \multicolumn{2}{c|}{Gigaword} &  \multicolumn{2}{c|}{Wikihow} \\
     \hline
     & Coverage & Faitfulness & Coverage & Faithfulness \\ 
     \hline
     Baseline   & 76.12  &  83.33  & 82.76  &  86.94   \\ 
     \hline
     \hline
     Loss Truncation & 79.55  &  87.17  &  84.93  &  87.84\\ 
     \hline
     \hline
     DAE & 78.23 & 86.33 & 84.15 & 88.83 \\
     \hline
     \hline
     Selector-ROC (Ours) &  64.58 &  84.17  & 78.67  &  87.84  \\
    \hline
    \hline
    Selector-F$_\beta$ (Ours)  &  &  &  & \\ 
    \hline
    $\beta$ &  &  &  & \\
    \hline
     0.5 & 54.77 & 76.83 & 64.24 & 79.82 \\
    \hline
    0.4 & 59.79 & 81.67 & 67.81  & 81.71 \\
    \hline
    0.3 & 60.72 & 82.00 & 68.53 & 83.15  \\ 
    \hline
    0.2 & 68.38 & 86.00 & 78.67 & 87.84  \\
    \hline
    0.1 & 79.92 & 88.00 & 84.72 & 89.19 \\ 
    \hline
    \end{tabular}
    \caption{Coverage and faithfulness scores for the baselines and our proposed methods. We show that with our method we are able to get models that are both more faithful and more abstractive than the baseline.}
    \label{tab:results}
\end{table*}

\noindent{\textbf{baseline \texttt{+} faithfulness (bf).}} We use the output from the baseline model if it is faithful (i.e.\ at least two out of three annotators agree that the output is faithful).
If the baseline output is not faithful, we select the output from the quartile model that is more extractive than the baseline to see whether we can have a similar coverage as the baseline but preserve faithfulness. 

\noindent{\textbf{baseline \texttt{+} faithfulness-extractiveness (bfe).}} This oracle system behaves similar to the one described above when the baseline output is unfaithful. However, rather than always selecting the baseline output when it is faithful, we pick the output from the quartile model that is more abstractive than the baseline whenever it is also faithful according to human judgement.

\noindent{\textbf{quartile \texttt{+} faithfulness-extractiveness (qfe).}}  Amongst the outputs of all four quartile models, we pick the most faithful output with the highest level of abstractiveness to understand whether it is possible to generate abstractive output while remaining faithful.

\noindent{\textbf{Analysis.}} Table \ref{tab:oracle_experiments} shows the coverage and faithfulness of the baseline and each of these oracles for Gigaword and Wikihow. We observe that it is possible to be more faithful than the baseline at a similar level of abstractiveness (bf). Furthermore, we can be more abstractive than the baseline while being more faithful (bfe). Selecting the most faithful and abstractive output from the quartile models achieves a really high faithfulness score 
($\approx$98\%) while having significantly less coverage than the baseline. This oracle analysis suggests that it should be possible to build models that can mitigate the faithfulness-abstractiveness trade-off by controlling the level of extractiveness.
Given this, we further explore whether we can learn a selector that is capable of doing this selection automatically to mitigate the faithfulness-abstractiveness trade-off. 

\begin{figure*}[t]
\centering
\begin{subfigure}{.5\textwidth}
  \centering
  \includegraphics[width=1\linewidth]{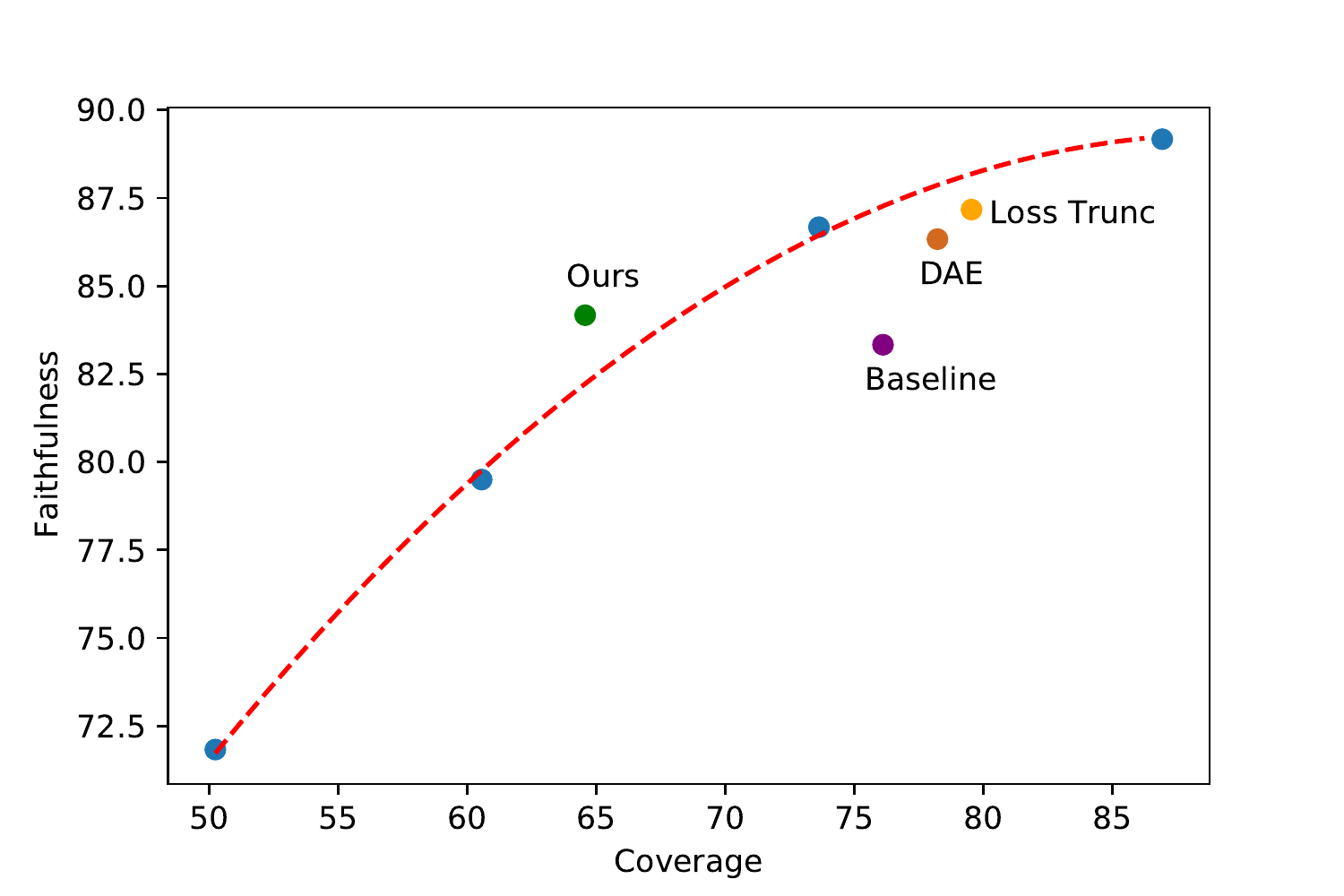}
  \caption{Selector-ROC and the baseline trade-off on \textbf{Gigaword}.}
  \label{fig:tradeoff_gigaword_selector_roc}
\end{subfigure}%
\begin{subfigure}{.5\textwidth}
  \centering
  \includegraphics[width=1\linewidth]{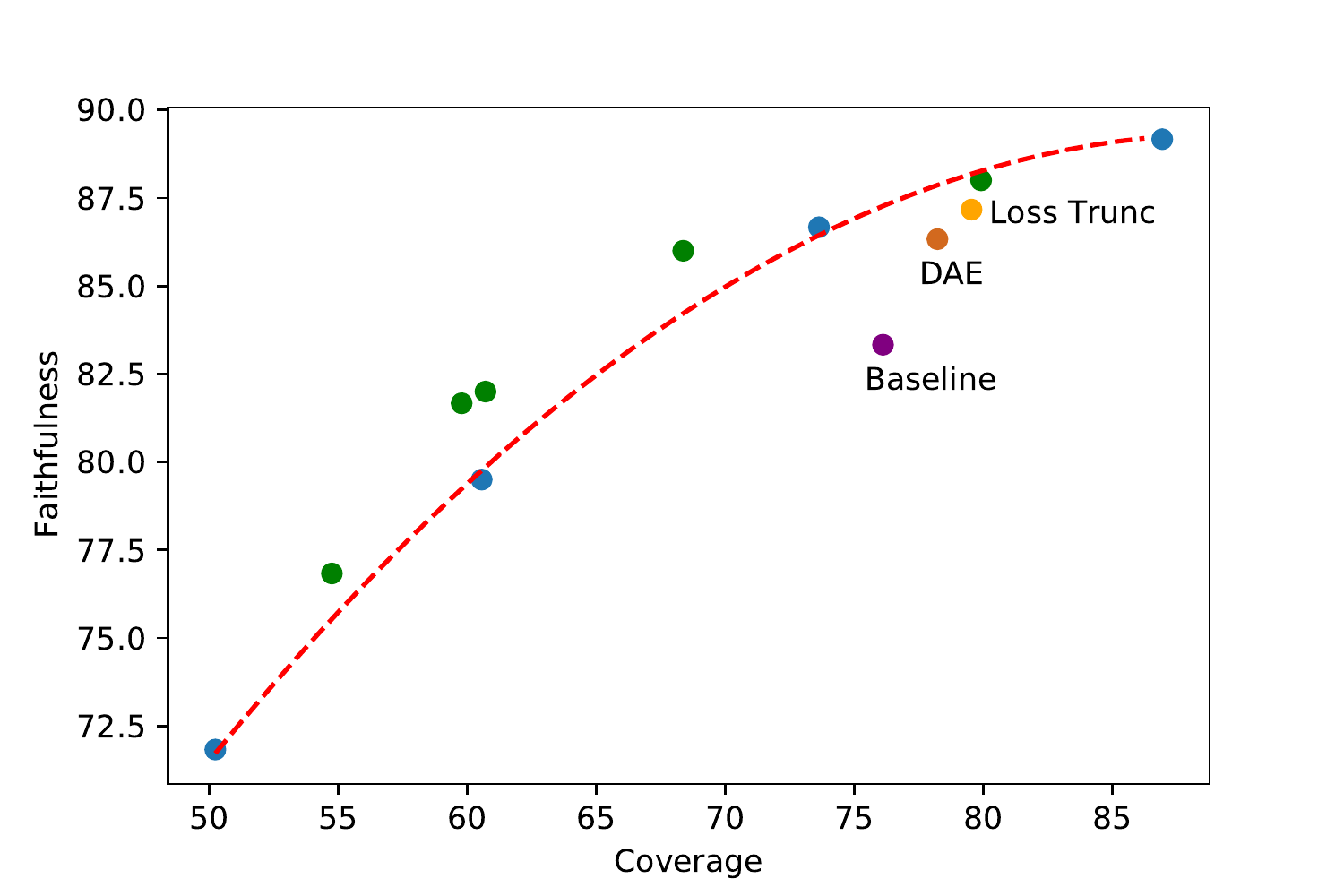}
  \caption{Selector-F$_\beta$ and the baseline trade-off on  \textbf{Gigaword}.}
  \label{fig:tradeoff_gigaword_selector_fbeta}
\end{subfigure}
\begin{subfigure}{.5\textwidth}
  \centering
  \includegraphics[width=1\linewidth]{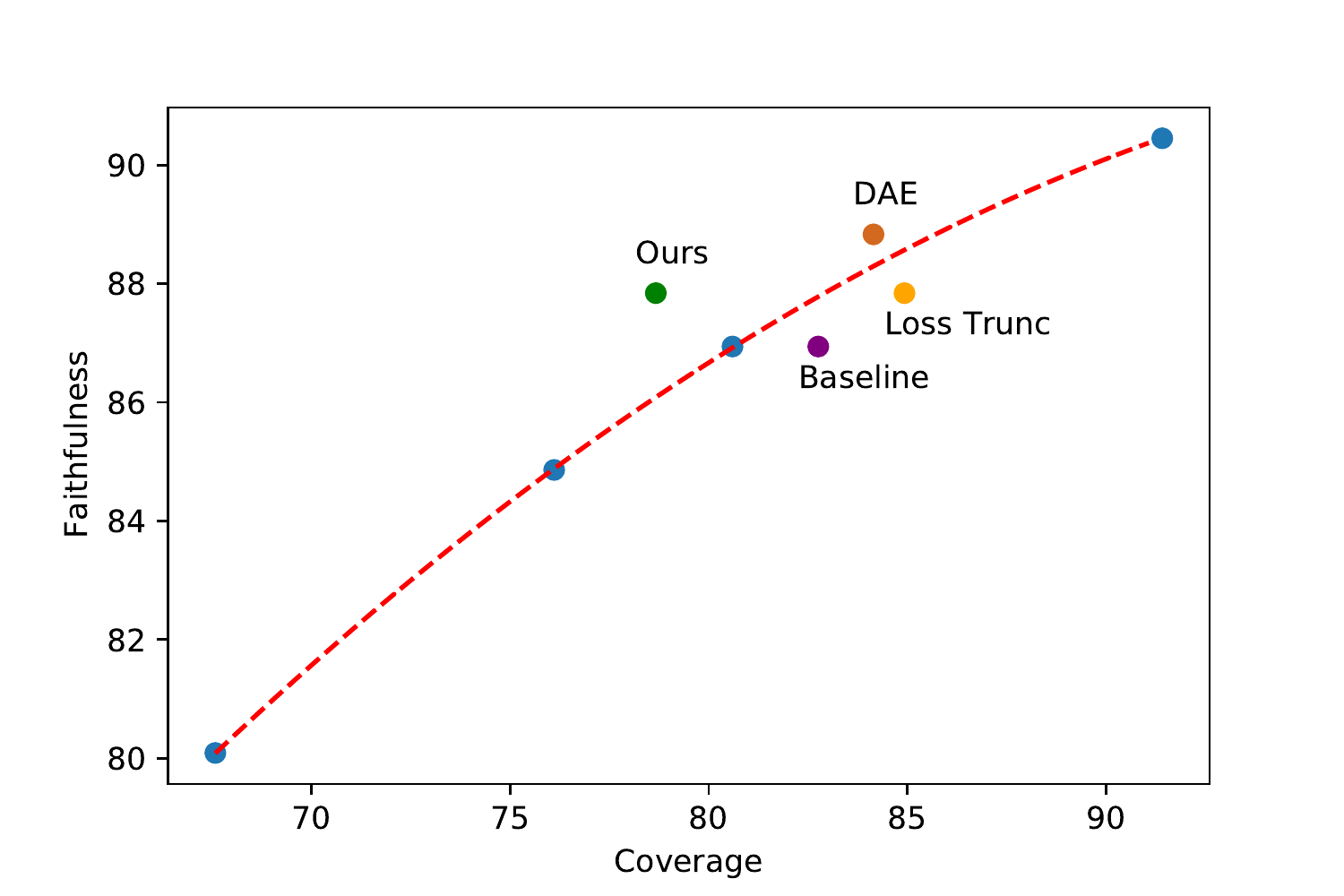}
  \caption{Selector-ROC and the baseline trade-off on  \textbf{Wikihow}.}
  \label{fig:tradeoff_wikihow_selector_roc}
\end{subfigure}%
\begin{subfigure}{.5\textwidth}
  \centering
  \includegraphics[width=1\linewidth]{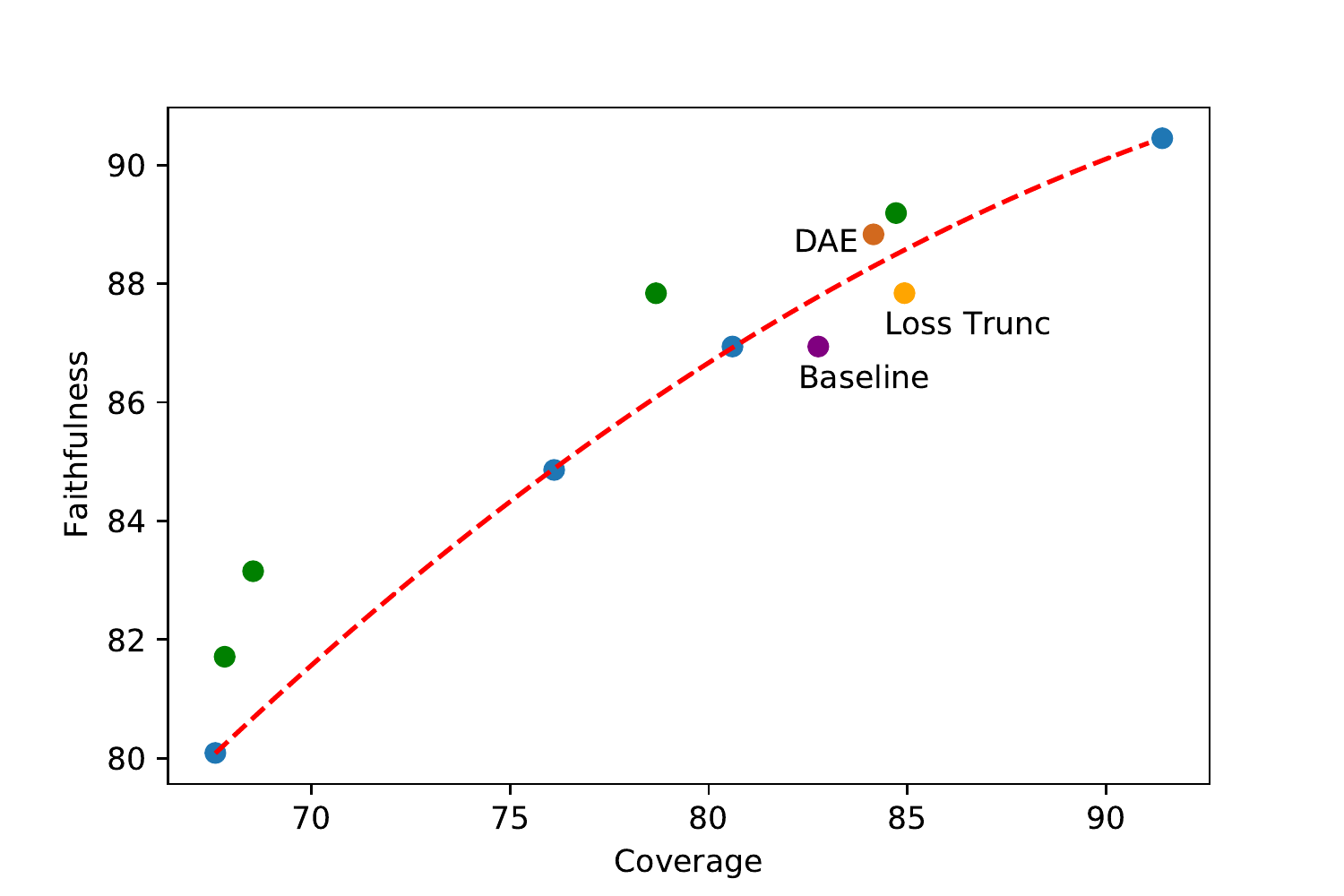}
  \caption{Selector-F$_\beta$ and the baseline trade-off on  \textbf{Wikihow}.}
  \label{fig:tradeoff_wikihow_selector_fbeta}
\end{subfigure}
\caption{Faithfulness-Abstractiveness trade-off curves. The blue dots represent the quartile models used to generate the curve. The purple dot corresponds to the baseline. DAE and Loss Truncation are depicted by the brown and orange dots respectively. The green dots correspond to our proposed systems.}
\label{fig:tradeoff_curves}
\end{figure*}

\subsection{Loss Truncation}

\newcite{kang-hashimoto-2020-improved} have proposed a method to adaptively remove high loss examples to optimize the distinguishability of samples from the model and the reference. They have shown that the samples generated by this Loss Truncation model achieves higher factuality ratings compared to the baseline methods. We study this method to understand where it lies in terms of faithfulness-abstractiveness trade-off and whether it can achieve a improved \textit{effective faithfulness} over the \textit{control}.

\subsection{Dependency Arc Entailment (DAE)}
\newcite{goyal-durrett-2020-evaluating} have proposed a factuality evaluation metric (DAE) that evaluates whether each dependency arc in the generated output is consistent with the input. They show that their proposed metric works better than existing factuality metrics, while also being able to localize the parts of the generated output that are non-factual. \newcite{goyal2020annotating} take advantage of DAE's ability to localize factuality errors, and train a summarization model only on the subset of tokens that is deemed factual according to the DAE metric. We follow their methodology to train summarization models, and assess them using our evaluation framework.

\subsection{Selector Model}

We aim to understand whether we can build a model that achieves a better \textit{effective faithfulness} than Loss Truncation. We propose a selector that can identify the most abstractive but faithful output to improve this trade-off. 
We first generate four possible candidate summaries using the quartile models for each example in the validation set. This results in outputs with varying levels of extractiveness.
For our selector, we fine-tune a FactCC model \cite{kryscinski-etal-2020-evaluating} on the data we collected to generate the trade-off curve, using 10-fold cross validation, to assign faithfulness scores to the generated summaries (in the test folds).\footnote{We collected annotations for 200 articles for each of the quartile models.}
In addition, we learn a threshold for the faithfulness score that maximizes the area under the ROC curve \textbf{(Selector-ROC)} (also using 10-fold cross validation). For each example in the test fold, we select the most abstractive candidate (amongst the four possible candidates from the quartile models) that is considered faithful according to the fintuned FactCC model (i.e. the faithfulness score is above the tuned threshold). 

Instead of maximizing for the area under the ROC curve, we can also tune the faithfulness threshold to maximize F$_\beta$ scores (\textbf{Selector-F$_\beta$}). Using F$_\beta$ score with $\beta<1$ allows us to assign a higher weight to the precision of our selector which would result in outputs with higher coverage and faithfulness. 

We find that the fine-tuning FactCC is important since the pre-trained FactCC model is trained on a different dataset and does not transfer well to our setttings. This is consistent with the findings of \newcite{goyal2020annotating}.

\begin{table*}[ht]
    \centering
    \begin{tabular}{|p{0.18\textwidth}|p{0.73\textwidth}|}
    \hline
    Article &  \begin{minipage}[t]{1.5\columnwidth}%
    If applicable, the description of any people who take part in your study should be extremely thorough. Each person should be identifiable within the research. Further, how people join and leave the study should be noted. If people were selected at random, or if they were family members, is important to the study. Be sure to consider various ethical concerns (e.g. risk and consent of participants) if people are involved in your research.
 \vspace{1mm}
\end{minipage}\tabularnewline 

    \hline
    Baseline & {\color{green}Describe} who is involved in the study. \\ 
    \hline
    DAE & {\color{green}Identify} the people who take part in the study.\\ 
    \hline
    Loss Truncation & {\color{green}Describe} people who take part in your study. \\ 
    \hline
    Selector-ROC (Ours) & {\color{green}Describe all} participants {\color{green}thoroughly} and {\color{green}with care}. \\ 
    \hline
    \hline
    Article &  \begin{minipage}[t]{1.5\columnwidth}%
    Because diarrhea frequently causes dehydration, it is crucial that patients with IBD remain hydrated. Drink at least 8 glasses of water every day (or 64 oz). Foods that have a high water content (like watermelon) can also count toward this minimum. If you have a severe attack of diarrhea, you are likely to lose electrolytes. In these cases, you might need to consume beverages such as Pedialyte or Gatorade to help replenish them [TRUNCATED] ...
    
 \vspace{1mm}
\end{minipage}\tabularnewline 

    \hline
    Baseline &  Drink {\color{green}plenty} of water to {\color{green}stay} hydrated. \\ 
    \hline
    Loss Truncation & Drink {\color{green}plenty} of water. \\ 
    \hline
    DAE & Drink {\color{green}plenty} of water to {\color{green}stay} hydrated.\\
    \hline
    Selector-ROC (Ours) & Drink {\color{green}plenty} of {\color{green}fluids} to {\color{green}stay} hydrated.\\ 
    \hline
    \end{tabular}
    \caption{Example summaries generated by the baseline, Loss Truncation and the selector model.}
    \label{tab:example_summaries_models}
\end{table*}

\subsection{Results}

Table \ref{tab:results} shows the coverage and faithfulness results for the baseline, Loss Truncation, DAE, and the selectors. We observe that as we use smaller values for $\beta$ for Selector-F$_\beta$, we get more extractive and more faithful outputs. This allows us to have a trade-off between faithfulness and abstractiveness. Moreover, with both Selector-ROC and Selector-F$_\beta$, we 
produce output with less coverage but higher faithfulness scores than the baseline. For Wikihow, Selector-ROC produces outputs with lower coverage but similar faithfulness scores to Loss Truncation. We can further obtain a higher faithfulness score at a similar coverage level as DAE and Loss truncation with Selector-F$_\beta$ with $\beta = 0.1$. For Gigaword, Select-ROC produces output with significantly lower coverage than Loss Truncation and DAE. Selector-F$_\beta$ produces output with similar coverage to Loss Truncation with a higher faithfulness score ($\beta = 0.1$). 

It is important to understand whether models improve faithfulness by simply being more extractive or if they are able to improve \textit{effective faithfulness}. In order to understand this, we  
measure whether the models get improvement in faithfulness over the \textit{control} operating at the same level of extractiveness. In Figure \ref{fig:tradeoff_curves}, we plot the faithfulness-abstractiveness curve with the faithfulness and abstractiveness of the quartile models. If a model lies above this curve, it improves the \textit{effective faithfulness}. If the model is below this curve, it is not able to improve the \textit{effective faithfulness} and it has a worse trade-off than the \textit{control} operating at the same level of extractiveness. 

For both Gigaword and Wikihow, Selector-ROC lies above the curve improving this trade-off. 
However, both the baseline and Loss Truncation models get worse trade-off than the \textit{control} operating at the same level of extractiveness. Similarly, we can obtain several models that lie above the curve for both Gigaword and Wikihow using Selector-F$_\beta$. The selector approach allows us to get better \textit{effective faithfulness} at different points in the abstractiveness-extractiveness spectrum. The DAE based model is able to improve \textit{effective faithfulness} on the Wikihow dataset, but not on the Gigaword dataset, indicating that the improvements are not consistent across datasets. 
Table \ref{tab:example_summaries_models} shows example summaries generated by the baseline, Loss Truncation, DAE and the Selector-ROC models. We observe that selector model is able to generate summaries that are faithful to the original article while having more novel words and phrases in the generated summaries.

%% file: related_work.tex
\section{Related Work}
There has been a lot of recent work in abstractive summarization showing that state-of-the-art systems suffer from generating inconsistent information with respect to the source article, despite their improved success in producing fluent summaries \cite{falke-etal-2019-ranking,  lux-etal-2020-truth, wilber-etal-2021-point}. Since word-overlap based metrics such as ROUGE have low correlation with human scores of faithfulness \cite{kryscinski-etal-2019-neural, fabbri2020summeval}, there has been 
significant effort to develop automated metrics that can detect such errors \cite{zhou-etal-2021-detecting, gabriel2021go-figure,pagnoni-etal-2021-understanding}. For example, \newcite{falke-etal-2019-ranking}, \newcite{maynez-etal-2020-faithfulness} and \newcite{ goyal-durrett-2020-evaluating} have proposed to assess faithfulness using entailment models, where a faithful summary should be assigned a high entailment score with respect to the original article. \newcite{kryscinski-etal-2020-evaluating}  presented FactCC, a weakly-supervised BERT-based entailment model, by augmenting the dataset with artificial faithfulness errors. \newcite{durmus-etal-2020-feqa} and \newcite{wang-etal-2020-asking} proposed question-answering based evaluation frameworks by automatically generating questions from the generated summary, and comparing the corresponding answers from both the source and the generated summary in order assess information consistency. Furthermore, several benchmarks have been proposed to evaluate the strengths and weaknesses of these evaluation metric \cite{gabriel2021go-figure,pagnoni2021understanding}. 

Previous studies in faithfulness evaluation, however, has not accounted for the effect of extractiveness of the output summaries. As we show in this study, the extractiveness of the output is correlated with the faithfulness scores assigned by these automated metrics. Therefore, it is not clear whether the models with higher scores are better at abstraction, or extract more from the source article. We suggest that we need to account for this confounding factor
in order to assess the real progress in building models that are better at abstraction. We note that there is concurrent work that also argues for accounting for extractiveness in assessing the faithfulness of models \cite{dreyer2021analyzing}, however, unlike our work, they do they do not propose any mitigation for the faithfulness-abstractiveness trade-off.

Improving faithfulness of summarization systems is essential for deploying these systems in real-world scenarios, as such recent work has studied methods to improve the faithfulness of  abstractive summarization systems \cite{matsumaru-etal-2020-improving, zhao-etal-2020-reducing, dong-etal-2020-multi, goyal2020annotating, xu-etal-2020-understanding-neural, chen-etal-2021-improving,zhu-etal-2021-enhancing}. For example, \newcite{goyal2020annotating} train summarization systems by modifying the training objective to maximize the likelihood of the subset of summary tokens that are considered faithful according to their factuality detection model. \newcite{zhao-etal-2020-reducing} specifically target hallucination of quantities in generated summaries, and train a verification model that they use to re-rank summaries such that summaries containing quantities consistent with the source article are up-ranked. 
Although these methods have shown improvements over the compared baselines, unlike our work, they do not measure the effective faithfulness taking extractiveness of the generated outputs into account.

%% file: implications.tex
\section{Implications and Limitations}

Recent studies that propose methods to improve faithfulness evaluate progress by conducting human evaluation on generated summaries and check whether the faithfulness 
scores are higher for their proposed system as compared to their baselines. We show that there is a strong relationship between the extractiveness and faithfulness of generated outputs (i.e., more extractive outputs tend to be more faithful), and therefore we cannot simply disregard extractiveness in faithfulness evaluation.

We propose that we should instead be measuring \textit{effective faithfulness} and introduce a framework that takes into account the \textit{faithfulness-abstractiveness trade-off curve} that is generated by training \textit{control models} at different points in the abstractiveness spectrum. We demonstrate the importance of measuring \textit{effective faithfulness} by showing that recently proposed methods that improve faithfulness over the baseline fails to consistently improve over a simple \textit{control} operating at the same level of abstractiveness. 

We argue that measuring \textit{effective faithfulness} is important since our goal is to build abstractive, faithful summarization systems. If the objective was to optimize for faithfulness alone, we could do so by simply building more extractive systems (such as the Q4 model we trained above).
\paragraph{Limitations.} Note that this method relies on some diversity in the extractiveness of reference summaries, since we rely on sub-sampling to train models for the \textbf{control}. It is less likely to be effective for datasets with very little variation in the extractiveness of the generated summaries. However, in general, we see significantly more faithfulness problems for  datasets with higher diversity of abstractiveness. Therefore, we suggest to account for the faithfulness-abstractiveness trade-off for such datasets in future work. 

\section{Acknowledgements}
This work was partly supported by the Office of the Director of National Intelligence (ODNI), Intelligence Advanced Research Projects Activity (IARPA), via contract \#FA8650-17-C-9117, Samsung Advanced Institute of Technology (Next Generation Deep Learning: From Pattern Recognition to AI), and 
a collaborative grant from Amazon to the Columbia Center for Artificial Intelligence entitled ``Extremely Abstractive Summarization''. The views and conclusions contained herein are those of the authors and should not be interpreted as necessarily representing the official policies, either expressed or implied, of the funding agencies. We further thank the anonymous reviewers and the Stanford NLP group for their helpful feedback. 

%% file: appendix.tex
\appendix

\section{Data Statistics}\label{sec:appendix_data_stats}

Number of examples, source article length and target summary length for each quartile are shown in Table \ref{tab:data_stats}. To create the quartiles, we first compute the extractiveness ($e_x$) of the reference summary, for each training example $x$, and compute the 25th ($a$), 50th ($b$), and 75th ($c$) percentile of the extractiveness of the training data. The quartiles are then created as follows:
\begin{align*}
    q1 &= \{x \mid e_x \leq a\} \\
    q2 &= \{x \mid a < e_x \leq b\} \\
    q3 &= \{x \mid b < e_x \leq c\} \\
    q4 &= \{x \mid e_x > c\}
\end{align*}

Note that it is possible for there to be several points at the boundary, and therefore there are unequal number of examples in each quartile as shown in Table \ref{tab:data_stats}.  For Gigaword, the article and summary lengths are very similar for each of the quartiles. For Wikihow, we observe that the article length is longer and summary length is shorter for more extractive quartiles. 

\section{Human Annotation Details} \label{human_eval_appendix}

We follow a similar procedure as the prior work to collect human evaluations for faithfulness of the generated summaries \cite{fabbri2020summeval}. Given the source articles and generated summaries, we ask annotators to judge whether the generated summary is \textbf{supported} by the article. The output is supported by the article if all the information expressed by the output can also be inferred from the article. We ask annotators to ignore minor grammatical errors and focus on the information content of the generated summaries. Figure \ref{fig:example_human_eval} shows an example from our human evaluation. 

\textbf{Computing faithfulness scores.} We evaluate $200$ output summaries per system and each output is evaluated by $3$ annotators. We restricted the study to the annotators with a high acceptance rate ($\geq98\%$) and at least 500 HITs to ensure annotation quality.\footnote{We hired annotators from USA, UK and Australia. The data collection protocol was approved by IRB.} We follow prior work \cite{durmus-etal-2020-feqa} and take the percentage of annotators who judge the summary as faithful to be the faithfulness score of a summary. To get the faithfulness score for a system, we average the summary scores across all $200$ samples. 

\section{Model details}
For all summarization models, we finetune BART ($406$M parameters) on a single Nvidia A-100 GPU. Each model takes roughly $3$ hours to train to convergence. For the selector, we finetune FactCC, on a single Nvidia A-100 GPU, using $10$-Fold cross validation. Finetuning for the entire cross validation procedure takes roughly $15$ minutes. We used all artifacts according to the terms indicated in their respective licenses.\footnote{\href{https://github.com/pytorch/fairseq/blob/main/LICENSE}{BART license.}}\footnote{\href{https://github.com/salesforce/factCC/blob/master/LICENSE.txt}{FactCC license.}}